\documentclass[10pt,conference,letterpaper]{ieeeconf}
\usepackage{amsmath,graphicx}

\usepackage{caption}
\usepackage{multirow}
\usepackage{subcaption}
\usepackage{url}
\usepackage{tikz}
\IEEEoverridecommandlockouts

\def\checkmark{\tikz\fill[scale=0.4](0,.35) -- (.25,0) -- (1,.7) -- (.25,.15) -- cycle;}

\title{Detection and prediction of clopidogrel treatment failures \\ using longitudinal structured electronic health records}
\author{Samuel Kim, In Gu Sean Lee, Mijeong Irene Ban, and Jane Chiang \\ \\
Cipherome Inc., San Jose, CA, U.S.A.\\
{\tt\small \{sam.kim, sean.lee, irene.ban, jane.chiang\}@cipherome.com}%
}

\begin{document}
%
\maketitle
\begin{abstract}
We propose machine learning algorithms to automatically detect and predict clopidogrel treatment failure using longitudinal structured electronic health records (EHR). By drawing analogies between natural language and structured EHR, we introduce various machine learning algorithms used in natural language processing (NLP) applications to build models for treatment failure detection and prediction. In this regard, we generated a cohort of patients with clopidogrel prescriptions from UK Biobank and annotated if the patients had treatment failure events within one year of the first clopidogrel prescription; out of 502,527 patients, 1,824 patients were identified as treatment failure cases, and 6,859 patients were considered as control cases. From the dataset, we gathered diagnoses, prescriptions, and procedure records together per patient and organized them into visits with the same date to build models. The models were built for two different tasks, i.e., detection and prediction, and the experimental results showed that time series models outperform bag-of-words approaches in both tasks. In particular, a Transformer-based model, namely BERT, could reach 0.928 AUC in detection tasks and 0.729 AUC in prediction tasks. BERT also showed competence over other time series models when there is not enough training data, because it leverages the pre-training procedure using large unlabeled data.

\end{abstract}

\section{Introduction}
\label{sec:intro}

Adverse drug reactions (ADRs) and treatment failures (TF), both of which can occur due to inadequate concentrations of medications in the system, significantly impact patients’ lives and are serious burdens to the healthcare system. ADRs, which may occur from excess active substance in the system, are the fourth leading cause of death behind cardiovascular diseases. Treatment failures may occur when therapeutic doses are not reached. Subsequent hospitalizations may significantly increase healthcare costs~\cite{LeadingCausesDeath2005}. The variable presentation makes identifying drug responses difficult, and the impact is likely underrepresented~\cite{ADRUnderReporting2006}. The multi-factorial nature of drug responses adds to the complexity; patient’s age, renal or liver function, comorbidities, comedications, lifestyle, and genetic predispositions are all important factors affecting drug responses~\cite{SignalADR2008,PharmacogeneticsADR2000}.

Among multitude of drugs, we studied clopidogrel, a P2Y12 inhibitor, one of the most widely prescribed drug in the US. Its use with aspirin is commonly referred to as dual antiplatelet therapy (DAPT), and is the standard of care following stent placement for cardiovascular disease. However, various drug responses to this drug has been reported in several studies raising concerns for its safety~\cite{ResistanceAntiPLT2007}. High concentration of active metabolite can cause bleeding, while low concentration of active metabolite can cause recurrent thrombosis, both of which can lead to medical emergencies. There is a clear need for a screening tool to better predict which patients may be at risk for such events.

Therefore, we focused on building machine learning models to automatically detect and predict the treatment failures on clopidogrel. More specifically, we used structured electronic health records (EHR) which contain information on many of the risk factors aforementioned. Easier access to medical records have enabled numerous applications using machine learning approaches. We used only structured EHR, in particular data generated by clinical coding schemes. Clinical episodes are coded by clinical experts to describe patients' health condition and treatments. Codes include disease diagnoses, procedures, and medicine prescriptions along with the dates. 

As structured EHR is inherently sequential, many researchers have applied time series modeling strategies. For example, Choi {\em et al.} proposed a recurrent neural network (RNN) architecture to detect heart failures~\cite{choiRETAINInterpretablePredictive2017}. The authors applied an attention mechanism on grouped diagnostic codes in a reverse time order and showed clinically interpretable results. Li {\em et al.} proposed BEHRT to learn patterns of diagnostic codes and used the model to predict unseen diagnoses~\cite{rasmyMedBERTPretrainedContextualized2021}. Although the original codes were generated in Read codes~\cite{readcode} and ICD-10 codes~\cite{icd10}, they were mapped into Caliber codes~\cite{mbbsChronologicalMap3082019} so that the vocabulary size became 301. Shang {\em et al.} proposed a combined architecture of Graph Neural Networks (GNNs)~\cite{wuComprehensiveSurveyGraph2021} and Bidirectional Encoder Representations from Transformer (BERT)~\cite{devlinBERTPretrainingDeep2019} for medication recommendation~\cite{shangPretrainingGraphAugmented2019}. The authors used the GNNs to represent ontology embeddings and the BERT to generate visit embeddings for downstream applications. 
Ramsy {\em et al.} proposed Med-BERT to build a disease classification such as heart failure among patients with diabetes (DHF) and pancreatic cancer (PaCa)~\cite{liBEHRTTransformerElectronic2020}. The model was trained using ICD-9/10 codes as tokens directly and fined-tuned for disease classification. 



In this paper, we investigated various time series modeling approaches to model the longitudinal structured EHR toward treatment failures detection and prediction. We hypothesized that the dynamic nature of medical records needs to be modeled. In that regard, we annotated clopidogrel treatment failures using publicly available data and built prediction models and detection models. Although previous related works are somewhat limited by only using diagnosis information, we proposed to use all possible data in addition to diagnoses, i.e., procedures and medication prescriptions. We hypothesized that they are complementary to represent patients' medical condition and treatments and modeling the interactions of them would provide better discriminating power. 


\section{Data}
\label{sec:data}
\subsection{Dataset}
We used the UK Biobank which consists of data collected from 502,527 participants ~\cite{UKB2015}. Volunteers aged 40 to 70 were recruited from England, Scotland, and Wales and invited to assessment centers between 2006 and 2010. Data collected during the visits include biosamples, physical examination measurements, and questionnaire answers followed by interviews. Genomic data, both sequencing and genotyping, have been generated using the biosamples collected as well. Also, an extensive and comprehensive medical history data have also been made available through hospital in patient and primary care data from external data sources. 

We extracted prescriptions, diagnoses and procedure records, along with dates, for all participants from the UK Biobank. All prescription records were from general practitioner (GP) data coded in Read and British National Formulary (BNF) depending on the data supplier. Diagnoses and procedure records were from hospital in-patient records only and coded in ICD-9/10 and OPCS-3/4 respectively.

\subsection{Annotation} 
\label{subsec:dataprep}

Treatment failure (TF) was defined as having a TF event within one year of the very first clopidogrel prescription. TF events include ischemic stroke, myocardial infarct, stent thrombosis and recurrent thrombosis or stenting, all of which were identified using ICD-9/10 and OPCS-3/4 codes. Clopidogrel prescriptions were identified using the substance or brand names of clopidogrel or respective read codes. 

Subjects with events occurring within 7 days of the first prescription were excluded as it is unclear whether those events were associated with clopidogrel. The visit had to be through the emergency room to be valid in order to exclude follow up visits from previous events. From the dataset, we found 9,867 subjects with clopidogrel prescriptions. Among them, we labeled 1,824 patients as treatment failure cases and 6,859 as control cases; 1,184 subjects were excluded due to data inconsistencies or ambiguities.  


\section{Methodologies}
\subsection{Data Processing}


We gathered various codes together per subject and organized them into visits with the same date. As a result, a subject has multiple visits and a visit has multiple codes. Table~\ref{table:demo_statistics} shows the statistics of the codes and the visits.

\begin{table}[!b]
  \centering
  \caption{Statistics of structured EHR data in UKBiobank.}
    \begin{tabular}{c|r}
    \hline
     \# of subjects & 502,527     \\
     \# of subjects with 1+ visits & 465,506 \\
     \# of unique visits & 22,063,417 \\
     \# of unique codes & 31,589 \\
     Average \# of visits per subject (max/med/min) & 58.3~(2805~/~13~/~1) \\
     Average \# of codes per visit (max/med/min) & 2.7~(61~/~2~/~1) \\
    \hline
    \end{tabular}%
  \label{table:demo_statistics}%
\end{table}%

\begin{figure}[b!]
  \centering
	  \includegraphics[width=\linewidth]{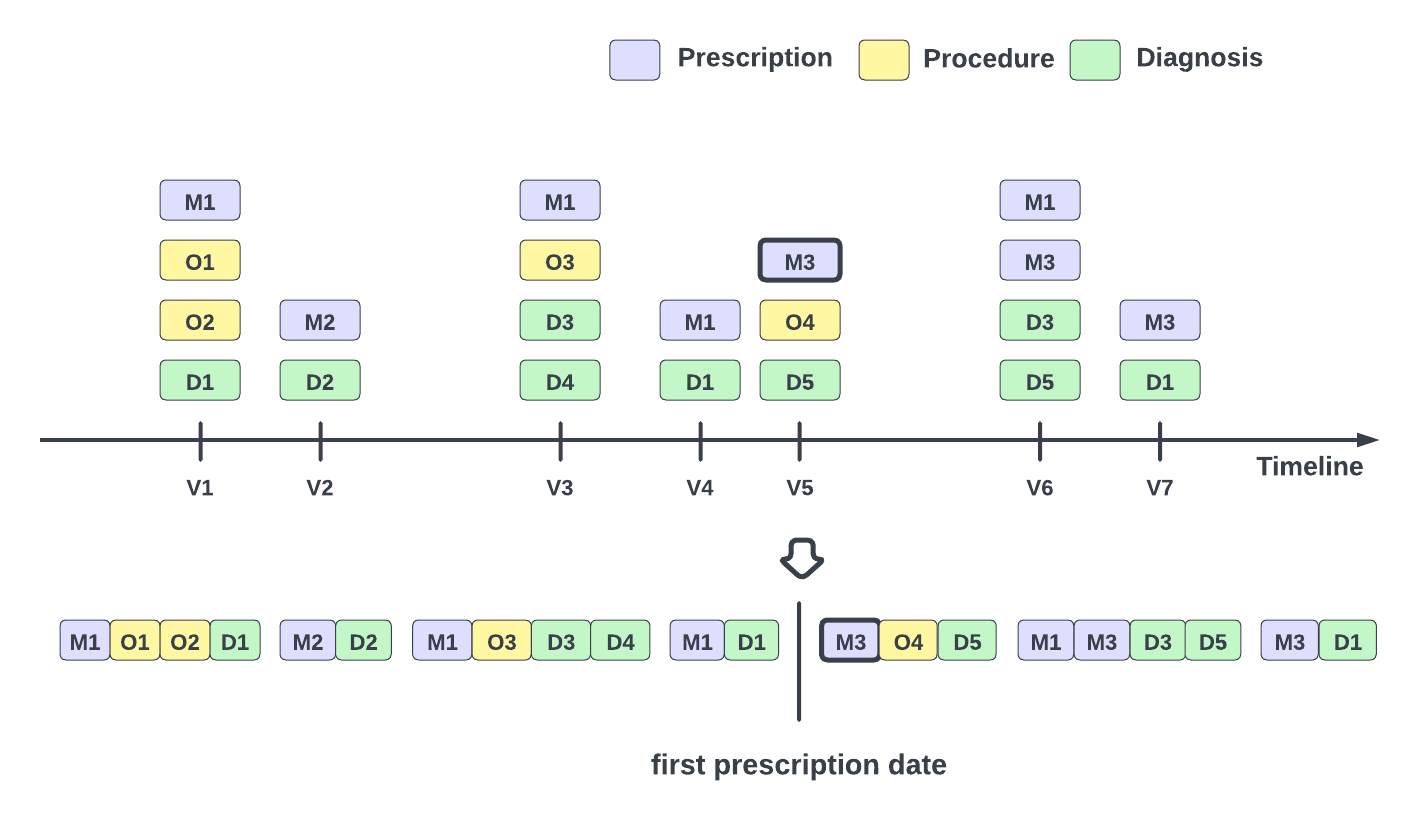}
	  \caption{Example of a patient's longitudinal structured EHR on timeline and how we process them. }
\label{fig:ehr_timeline}
\end{figure}

Note that the number of codes per visit is relatively small compared to the number of words per sentence in natural language processing (NLP) applications. Also, the codes within in a visit do not have obvious ordinality. On the other hand, the number of visits per patient is fairly large and their order is less ambiguous due to timestamps. Therefore, we flattened the codes of the visits and concatenated them across all visits so that a patient has a ordered sequence of codes as illustrated in Fig.~\ref{fig:ehr_timeline}.

\subsection{Model Architectures}
\label{subsec:pretrain}
We used various machine learning algorithms including gated recurrent neural network (GRU)~\cite{gru}, and long short-term memory (LSTM)~\cite{lstm}, and BERT~\cite{devlinBERTPretrainingDeep2019} for time series modeling. We also used random forest (RF) and logistic regression (LR) with bag-of-words approaches as baselines.

For recurrent-based neural network approaches such GRU and LSTM, we empirically chose one hidden layer of GRU or LSTM with 768 units. For BERT models, we used the original architecture in~\cite{wolf-etal-2020-transformers} for simplicity. In both approaches, only the most recent 512 codes per subject were used if there were more. To build a pre-trained model for BERT, we used the data from the subjects that were not annotated with treatment failures. Then, we fine-tuned the pre-trained model using the labeled data. We replaced the top layer of the pre-trained model with a fully connected layer with one node with binary cross-entropy as a loss function.

We also built a tokenizer that converts individual clinical codes into unique integer values to feed into machine learning algorithms. The tokenizer is built using all the codes together unless noted otherwise.

\subsection{Experimental Setup}

We designed two experimental tasks: 1) detecting treatment failures within a year after the first prescription of clopidogrel and 2) predicting treatment failures within a year after the first prescription of clopidogrel. The main difference of these two tasks is the direction of point of views on the timeline. The former is looking back from one year after the first prescription if there have been any treatment failures, while the latter is looking forward at the first prescription date to predict if there will be any treatment failures. From the data perspectives, the former has an access to the all data until one year after the first prescription date while the latter only has access the data before the first prescription date (see Fig.~\ref{fig:ehr_timeline} for an example).


We randomly split the labeled data into training and testing data with 80/20 ratio. This split was used across all experiments for fair comparisons. Twenty percent of the training data was used as validation data if necessary.  We used receiver operating characteristic (ROC) curves and their areas under curve (AUC) to measure the performance of each experiment.

\section{Results and Discussion}
\label{sec:results}

\begin{figure}[t!]
  \centering
  \begin{subfigure}{0.47\textwidth}
	  \includegraphics[width=\linewidth]{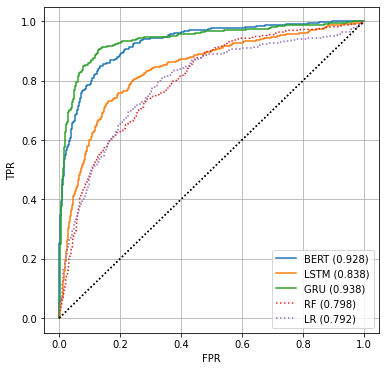}
	  \caption{Detection tasks}
  \end{subfigure}
  \begin{subfigure}{0.47\textwidth}
	  \includegraphics[width=\linewidth]{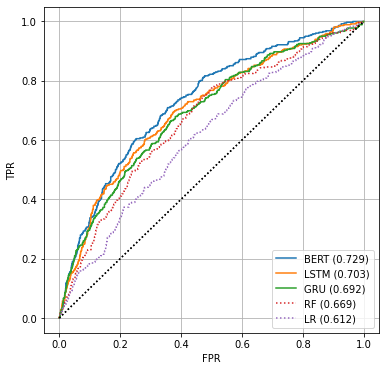}
	  \caption{Prediction tasks}
  \end{subfigure}
  \caption{Performance of the models for adverse drug reactions in terms of ROC and its AUC: (a) detection tasks and (b) prediction tasks.}
\label{fig:roc}
\end{figure}

Fig.~\ref{fig:roc}~(a) and (b) illustrates the performance of the models for the detection and prediction tasks. As shown in the figures, the results of detection tasks are better than the prediction tasks. This is reasonable because the detection tasks are to make decisions with more data after the prescription of the drug while the prediction tasks are to predict only with the data before the prescription. 

\begin{figure}[t!]
  \centering
  \begin{subfigure}{0.47\textwidth}
	  \includegraphics[width=\linewidth]{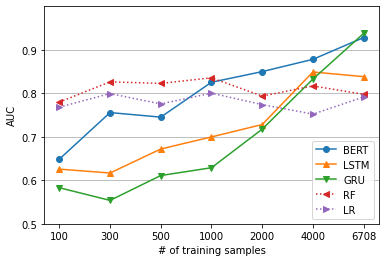}
	  \caption{Detection tasks}
  \end{subfigure}
  \begin{subfigure}{0.47\textwidth}
	  \includegraphics[width=\linewidth]{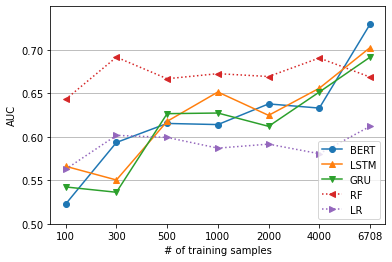}
	  \caption{Prediction tasks}
  \end{subfigure}
  \caption{Performance of the models for adverse drug reactions with respect to number of training samples in terms of AUC.}
\label{fig:amount}
\end{figure}

It is notable that using time series modeling strategies such as BERT, LSTM and GRU outperforms the ones using bag-of-words approaches such as RF and LR in both tasks. This vindicates that the dynamics in the longitudinal medical records convey significant information regarding drug responses. Note that the best model for detection tasks is GRU (AUC=0.938) rather than BERT (AUC=0.928). Although we initially expected BERT to be the best model as it used a large unlabeled dataset to build the pre-trained model and fine-tuned with the labeled data, it was somewhat difficult to observe benefits of the pre-trained model. 

\begin{table}[b!]
    \centering
    \caption{Performance of the models for adverse drug reactions with different types of data in terms of AUC. Checkmarks are the modalities that were used to build the corresponding model.}
    
    \begin{tabular}{c|c|c||c|c|c||c|c}
    \multicolumn{8}{c}{(a) Detection tasks } 
    \\ \hline
        Proc. & Diag. & Pres. & BERT & LSTM & GRU & RF & LR  \\ \hline \hline
        \checkmark & \checkmark & \checkmark & \bf 0.928 & \bf 0.838 & \bf 0.938 & 0.798 & 0.792  \\ \hline
        \checkmark & ~ & ~ & 0.716 & 0.805 & 0.801 & 0.779 & 0.793  \\ \hline
        ~ & \checkmark & ~ & 0.759 & 0.810 & 0.809 & 0.771 & 0.769  \\ \hline
        ~ & ~ & \checkmark & 0.721 & 0.723 & 0.722 & 0.725 & 0.628  \\ \hline
        \checkmark & \checkmark & ~ & 0.829 & 0.820 & 0.834 & \bf 0.810 & \bf 0.813  \\ \hline
        ~ & \checkmark & \checkmark & 0.877 & 0.822 & 0.882 & 0.794 & 0.776  \\ \hline
        \checkmark & ~ & \checkmark & 0.808 & 0.726 & 0.887 & 0.771 & 0.750 \\ \hline
    \end{tabular}
    
    \begin{tabular}{c|c|c||c|c|c||c|c}
    \\
    \multicolumn{8}{c}{(b) Prediction tasks }
    \\ \hline
        Proc. & Diag. & Pres. & BERT & LSTM & GRU & RF & LR  \\ \hline \hline
        \checkmark & \checkmark & \checkmark & \bf 0.729 & \bf 0.703 & \bf 0.692 & 0.669 & 0.612  \\ \hline
        \checkmark & ~ & ~ & 0.606 & 0.632 & 0.624 & 0.594 & 0.602  \\ \hline
        ~ & \checkmark & ~ & 0.604 & 0.623 & 0.613 & 0.643 & 0.628  \\ \hline
        ~ & ~ & \checkmark & 0.615 & 0.648 & 0.691 & \bf 0.669 & 0.563  \\ \hline
        \checkmark & \checkmark & ~ & 0.622 & 0.663 & 0.666 & 0.659 & \bf 0.640  \\ \hline
        ~ & \checkmark & \checkmark & 0.605 & 0.667 & 0.662 & 0.672 & 0.581  \\ \hline
        \checkmark & ~ & \checkmark & 0.615 & 0.679 & 0.664 & 0.675 & 0.600 \\ \hline
    \end{tabular}
    \label{table:ablation}
\end{table}

Therefore, we performed extra experiments with different training data sizes. We randomly chose a subset of training data with variable sizes to simulate the cases where there were not sufficient training data. Note that we kept the same test dataset across the experiments for fair comparison. Fig.~\ref{fig:amount} depicts AUCs of the models with respect to number of training samples. As shown in the figures, it is evident that the AUCs of sequential modeling approaches (solid lies) degrade as the number of training samples decreases while bag-of-words approaches (dotted lines) do not have clear patterns. This is partially because the number of parameters of sequential models are far greater than the ones of bag-of-words models so they suffer from lack of data. It is also notable that BERT outperforms other models with smaller training datasets particularly in detection tasks. We argue that BERT benefits from leveraging the pre-training procedure and quickly adopting their parameters with the labeled training data.

We also performed ablation experiments which enable us to analyze the impact of different types of medical records. For these experiments, we built models for individual modalities, i.e. procedures, diagnoses, and prescriptions, and also combinations. We also built separate tokenizers for each experiment. As shown in Table~\ref{table:ablation}, the models with all the modalities outperform the ones with partial information especially using sequential models. This confirms that modeling the interactions in multiple modalities in structured EHR provides better predictive power.

\section{Conclusions}
We introduced ways to model longitudinal structured EHR toward detecting and predicting treatment failures on clopidogrel. We applied time series modeling strategies that are commonly used in NLP applications. The experimental results showed that the algorithms that model time dynamics outperformed bag-of-words approaches and suggested to use pre-trained models when the amount of labeled data is small. It was also shown that using various types of medical records together improved the performance in detecting and predicting treatment failures. 

In future, we will investigate visit embeddings to represent all the codes within a visit to mitigate ordinality issues. We also want to incorporate with lab data such as glomerular filtration rate, liver function test, and hemoglobin levels, to improve model performance.

\section{Acknowledgement}
This research has been conducted using the UK Biobank Resource under Application Number 52031.

\bibliographystyle{IEEEbib}

{\bibliography{mybib}}

\end{document}